# Inception-Residual Block based Neural Network for Thermal Image Denoising

Seongmin Hwang, Gwanghyun Yu, Huy Toan Nguyen, Nazeer Shahid,
Doseong Sin, Jinyoung Kim and Seungyou Na
Chonnam National University
77 Yongbong-ro,
Buk-gu, Gwangju 61186, Korea
min6766e@gmail.com, sayney1004@naver.com, nguyenhuytoantn@gmail.com, nazeershahid12936@gmail.com,
saintds33@gmail.com, beyondi@jnu.ac.kr, syna@jnu.ac.kr

## Abstract

*Thermal cameras show noisy images due to their limited thermal resolution, especially for the scenes of a low temperature difference. In order to deal with a noise problem, this paper proposes a novel neural network architecture with repeatable denoising inception-residual blocks(DnIRB) for noise learning. Each DnIRB has two sub-blocks with difference receptive fields and one shortcut connection to prevent a vanishing gradient problem. The proposed approach is tested for thermal images. The experimental results indicate that the proposed approach shows the best SQNR performance and reasonable processing time compared with state-of-the-art denoising methods.*

1. Introduction

The noise in digital images occurs because of several reasons such as collection, coding, transmission and processing. Image denoising techniques are implemented in low-level image processing which has been an active topic in the computer vision field. The main purpose of the denoising technique is recovering a clean image x from a noisy image y (y = x + v, where x: clean image, v: noise).

For decades, several approaches have been used to handle noising problems such as Gaussian Smoothing [1], the anisotropic filtering [2, 3], the Total Variation minimization [4], wavelet thresholding methods [5, 6], and BM3D [7]. Nonlocal self-similarity (NSS) models (BM3D [7], LSSC [8], NCSR [9] and WNNM [10]) had achieved high performance. More specifically, Block Matching and 3D Collaborative Filtering (BM3D) was realized as the best approach in image noise reduction. Image prior-based methods, which recover original images from noisy images with no prior information, have good performance of removing noise in image. However, these approaches have some limitations as follows. First, image prior-based methods involve a complex optimization task leading to a high computational cost. Second, hand-designed parameters are difficult to select.

In the last ten years, a variety of discriminative learning methods were researched to overcome aforementioned problems. By implementing Multi-layer perceptron (MLP) [11] on GPU with a large dataset, Burger et al. was able to achieve good denoising performance. Schmidt et al. suggested cascade of shrinkage fields (CSF) which combine a random-field based model and an optimization algorithm [12]. Trainable nonlinear reaction diffusion (TNRD) is proposed to train parameters in the diffusion model based on the gradient descent inference steps [13].

Thanks to the development of the hardware technology especially in GPU, the deep learning method was recently adopted for removing noise in images. Particularly, Zhang et al. [14] introduced the CNN denoiser (DnCNN) to train parameters by residual learning. Byeongyong Ahn, and Nam Ik Cho proposed a block-matching convolutional neural network (BMCNN) [18], which combines the NSS prior and the CNN. These approaches were able to overcome problems of previous methods; however, they still required long processing time. There were many approaches based on deep learning, which were introduced to solve a noise problem for RGB images, but none of them were not intended for removing noise in the thermal images. In fact, once taking an image by the thermal camera in the environment in which the difference between the minimum and maximum temperatures is less than 2℃, random noise exists because of quantization.

In this paper, a novel neural network architecture with the repeatable denoising inception-residual block (DnIRB) for noise learning is proposed. Our proposed method attains high performance and fast processing time, compared to the previous methods.

This paper is organized as follows. First of all, the problem of the thermal image noise is defined in Section 2. The proposed model architecture with repeatable inception-residual blocks to explain residual learning is introduced in Section 3. The experimental set-up and results are presented in Section 4. Finally, the conclusion is provided in Section 5.

## 2. Problem definition

Since the image is obtained using the thermal camera, if the temperature difference among the objects in the scene is less than 2 degree Celsius, the noising problem appears due to the quantization of camera. The noise extractor based on BM3D model [7] is utilized to obtain the noise histogram from the original noise images. The comparison of noise pdf (probability density function) including Gaussian noise, Laplace noise and extracted thermal image noise is represented in Fig. 1. In each graph, the horizontal axis is the noise value and the vertical axis is the probability with noise.

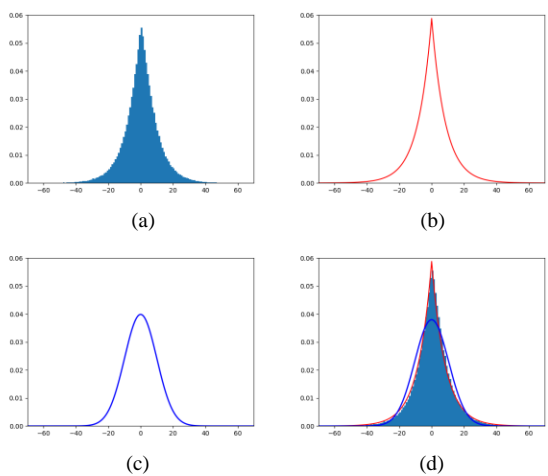

Figure 1: Comparison of Gaussian and Laplace PDF: (a) Extracted thermal noise distribution PDF; (b) Laplace PDF; (c) Gaussian PDF; (d) Comparison of each PDF

The thermal noise is depicted in Fig. 1 (a). From the figure, we are able to observe that the extracted thermal noise value is in range of -40 to 40. As shown in Fig. 1 (b), the Laplace noise has the same noise range. Conversely, Gaussian noise in Fig. 1 (c), represents a crucial different distribution range. In Fig. 1 (d), the extracted thermal noise distribution is overlapped with Laplace PDF (red color) and Gaussian PDF (blue color) to be obtained that Laplace PDF is more likely to approximate thermal noise. Based on these observations, the Laplace noise distribution is selected to train the denoising model. The experimental results will be shown in Section 4.

## 3. Proposed Method

In this section, the proposed model architecture based on the Denoising Inception-Residual block is presented. After that, the details of Repeatable Inception-Residual Block are carefully analyzed to show our contribution. Finally, Residual Learning is explained.

### 3.1. Model Architecture

An overview of the proposed method is depicted in Fig. 2. Our work has been inspired by the DnCNN network [14]. However, to improve the training speed, we proposed to apply Inception-ResNet [15] instead of normal convolutional layers in the middle of the network. At a low feature level, both a simple convolution layer and a complex convolution layer show the similar results [17]. Therefore, at the first and second layers, the 7×7 and 3×3×64 convolution filters are adopted respectively to obtain the 64 low-level feature map. The repeatable denoising inception-residual blocks (DnIRB) is then configured. This replacement significantly enhances the network performance. Finally, the single convolution filter of 3×3×64 is applied to get the noise output.

### 3.2. Repeatable Denoising Inception-Residual Block

Recently, the convolutional neural network achieved good performance in the image classification and recognition. Specially, the Inception-Resnet [15], which contains a variety of receptive fields and short-cut connections, showed remarkable results in both processing time and performance. The benefit of the Inception-Resnet is investigated in our proposed method to enhance denoising results of thermal images.

The architecture of the repeatable DnIRB is illustrated in Fig. 3. This block is divided into two sub-blocks. Firstly, the 1×1 convolution filter is applied for each sub-block based on the bottleneck process to reduce the number of channels from 64 to 32 in order to decrease the computational cost. One sub-block contains one 3×3 convolutional layer. Conversely, the other has two 3×3 convolutional layers as shown in Fig. 3. The local features of two sub-blocks with different sizes are collected and concatenated. The short-cut connection is directly applied between input and output to prevent the vanishing gradient in a deep neural network. The short-cut connection performs an identity mapping and its outputs are added to the concatenated layers. The short-cut connection does not add additional parameters and computational complexity [16].

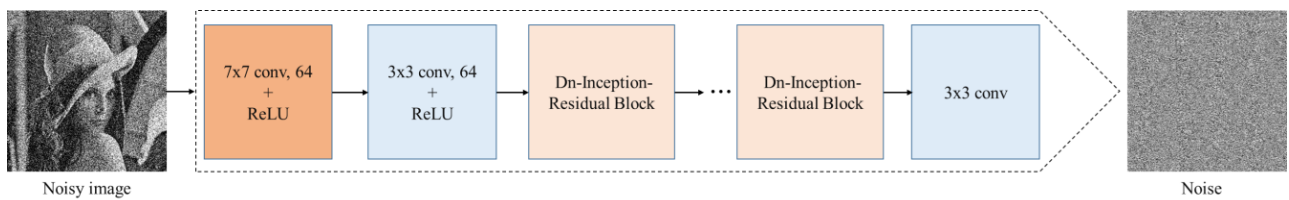

Figure 2: Proposed Model Architecture

### 3.3. Residual Learning

An input noisy image (y) is generated by the sum of clean image (x) and noise (v) (e.g. y = x + v).

The discriminative denoising model such as MLP [11] was trained to estimate clean image (D(y) = x). In contrast, the proposed approach is based on a residual learning method, which learns noise estimator (R(y) = v). The loss function of residual learning is defined in Equation (1).

$$L(\theta) = \frac{1}{n}\sum_{i=1}^{n}\|R(y_i;\theta) - n_i\|^2 \quad (1)$$

where, R is noise estimator with parameter θ, and n is noise. Generally, estimating clean images of noisy observation are more complex than training noise image learners [14].

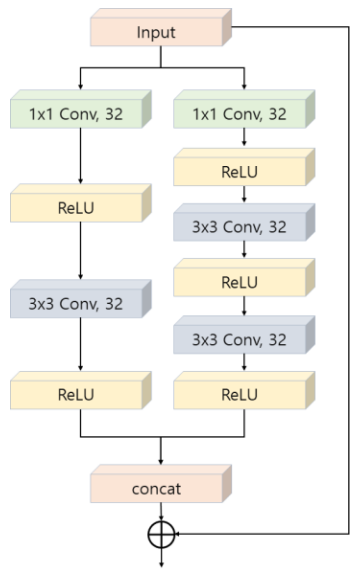

Figure 3: Repeatable denoising inception-residual block (DnIRB)

## 4. Experimental evaluation

### 4.1. Experimental set-up

We trained the proposed model by our own dataset with the 640×480 pixels resolution. The dataset was recorded at Chonnam National University by using the Argo-S thermal camera at 31fps. To increase the number of training samples, each image is separated to small patches with 40×40 size and 14 strides. After that, the data augmentation methods are applied including flip, scaling, and rotation. Finally, the total number of image patches are 1,044,548. We train the proposed model using TensorFlow by Nvidia Tesla K40c GPU. In the testing phase, the Intel(R) Core (TM) i7-4790 CPU 3.60Hz and 8GB RAM is used. We evaluate the denoising performance of the proposed method in comparison with the previous state-of-the-art methods in both accuracy and computational cost aspects using our own database.

### 4.2. Evaluation

In this section, we evaluate our proposal approach with other state-of-the-art methods by showing the critical novel results to confirm the effectiveness of our proposed method in various scenarios with different noises and impacts.

*Laplace Noise*

Table 1 shows the comparison of the average PSNR in the testing phase between the proposed method and other recent state-of-the art approaches with Laplace noise. The impact of a scale variable is also examined. Based on Table 1, we can observe that the proposed method obtains the best performance compared with other state-of-the-art methods with different scale variables up to 44.31dB at scale $b = 5$. The proposed method outperforms the benchmark-BM3D method by 0.49, 1.02, 1.48, 2.38 dB at scale values of 5, 7.5, 12.5 and 25, respectively. More notably, our method is better than the latest method based on DnCNN [14] 2.34dB at the scale value of 25. From Table 1, we can realize that as the scale value is increasing, the average PSNR (dB) is decreasing. This result relies on the fact that with higher scale values, the noise level is higher due to loosing information in scaling steps. The denoising results with Laplace noise ($b = 12.5$) are represented in Fig. 4.

*Table 1: The average PSNR(dB) of Laplace Noise*

| Scale | $b = 5$ | $b = 7.5$ | $b = 12.5$ | $b = 25$ |
|---|---|---|---|---|
| Noisy Image | 30.95 | 27.47 | 23.14 | 17.56 |
| BM3D[7] | 43.82 | 42.05 | 39.41 | 35.1 |
| WNNM[10] | 43.47 | 41.58 | 38.58 | 33.89 |
| MLP[11] | - | - | 38.26 | 33.38 |
| TNRD[13] | - | 41.52 | 38.93 | 34.44 |
| DnCNN[14] | 43.69 | 42.1 | 39.57 | 34.75 |
| Proposed (4-blocks) | **44.31** | **43.07** | **40.89** | **37.48** |

*Gaussian Noise*

Gaussian noise removing ability is also examined as shown in Table 2. Considering that most of the recent approaches for denoising were implemented by Gaussian noise, we trained our network model with Gaussian noise and investigated its performance. In this scenario, the noise level (σ) is varied from 10 to 50. The detailed experimental results are provided in Table 2. From Table 2, one can observe that with the denoising performance at all levels, the resulting DnIRB achieved the highest average PSNR (dB). The proposed method improves the base-line DnCNN and the benchmark BM3D by 1.96 dB and 1.98 dB at the noising level σ=50, respectively. From Table 2, the same conclusion which is drawn with Laplace noise is that denoising performance is better at the lower noise level. The illustration of denoising performance is provided in Fig. 5 with Gaussian noise at a level 25.

In addition, the noise distribution of thermal images was previously mentioned in Section 2. Most of the cases, where the thermal image noise has the same shape with Laplace noise which leads to the trained model, have the best performance with Laplace noise. Therefore, in later experiments, we only consider the Laplace denoising model.

*Table 2. The average PSNR(dB) of Gaussian Noise*

| Sigma | σ = 10 | σ = 15 | σ = 25 | σ = 50 |
|---|---|---|---|---|
| Noisy Image | 28.13 | 24.63 | 20.29 | 14.73 |
| BM3D[7] | 42.31 | 40.28 | 37.36 | 32.52 |
| WNNM[10] | 42.39 | 40.32 | 37.25 | 32.18 |
| MLP[11] | - | - | 37.74 | 33.84 |
| TNRD[13] | - | 40.08 | 37.33 | 32.22 |
| DnCNN[14] | 42.39 | 40.77 | 38.12 | 32.54 |
| **Proposed (4-blocks)** | **42.74** | **41.08** | **38.66** | **34.50** |

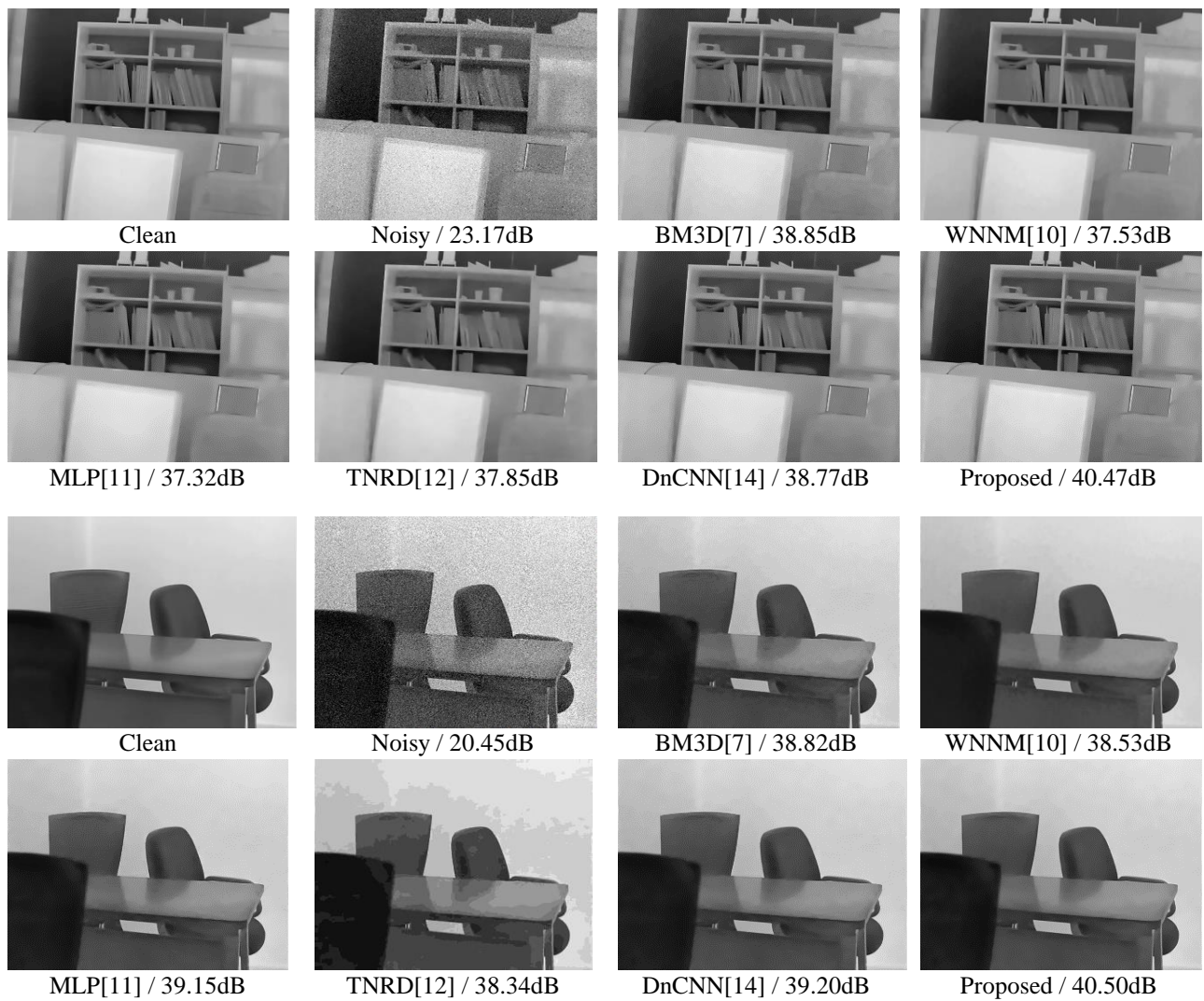

Figure 5: Denoising results with Gaussian noise level 25

*Impact of block number*

In order to investigate the influence of the number of blocks in a training mode, we trained a different model with various numbers of blocks including 2 blocks, 4 blocks, 8 blocks, and 16 blocks. The effects of the number of repeatable denoising inception-residual blocks (DnIRB) are investigated in Table 3. We have found that the increasing number of blocks from 2 to 4 blocks make a significant improvement of 0.63 dB with the processing time increase 0.5308seconds. However, from 4 blocks to 8 blocks, the performance only increases 0.36dB, but the processing time is 0.9763 seconds higher. In principle, we can choose a higher number of blocks, which can make the proposed model deeper and give better denoising performance. However, we practically need to consider the trade-off between the accuracy and the processing time. From our experiment, we recommend using the model with 2 blocks for this purpose.

*Computational Cost*

We implemented our training phase on Nvidia Tesla K40c GPU and testing phase on the Intel(R) Core(TM) i7-4790 CPU 3.60Hz and 8GB RAM. The CPU is used for testing to compare the processing time. We run our model with the two-block case. The computational cost is provided in Table 4. The proposed method is two times faster than DnCNN and has achieved the lowest computation cost. It is reasonable, because the Inception-Residual Block in each stage significantly reduces the computational cost. Conversely, WNNM [10] has an acceptable denoising result only with a very high computational cost, which makes it difficult to use in a real-time processing application.

In summary, our DnIRB outperforms the recent state-of-the-art methods in both processing time and accuracy. Our approach is able to be applied to a real-time denoising application using CPU.

The denoising result with original thermal image noise is provided in Fig. 6. Blue and orange colors represent blurred detail and faded edges respectively. Our proposed method shows the best results compared to other methods.

5. Conclusions

While taking images by thermal cameras within a distance close to target objects, we can observe noisy images. To remove noise from thermal images, we proposed the Deep Learning Based Denoising Model using the end-to-end mapping with Residual Learning. Our model contains the special unit, namely, repeatable denoising inception-residual blocks (DnIRB). DnIRB has two separate sub-blocks to obtain various receptive fields. To avoid a vanishing gradient, the direct short-cut connection between the input and the output is adopted. Our method is compared to various state-of-the-art approaches using our own thermal image dataset and the experimental results showed that the proposed approach attains the best PSNR with a small calculation amount. On the other hand, with the increase of repeatable DnIRB, the proposed method gets higher PSNR; however, the method requires more processing time. Therefore, the proper number of blocks should be selected considering the PSNR and time consuming for a given application. Our achievements can be employed in various applications to enhance not only thermal images but also visual images.

Acknowledgment

This research was supported by the MSIT(Ministry of Science and ICT), Korea, under the ITRC(Information Technology Research Center) support program (IITP-2018-2016-0-00314) supervised by the IITP(Institute for Information & communications Technology Promotion).

*Table 3. The average PSNR(dB) and Time according to number of blocks*

| Number of blocks | PSNR(dB) | Time(sec) |
| --- | --- | --- |
| 2-blocks | 40.26 | 0.7508 |
| 4-blocks | 40.89 | 1.2816 |
| 8-blocks | 41.25 | 2.2579 |
| 16-blocks | 41.55 | 4.2563 |

*Table 4. Comparison of processing time with PSNR*

| Method | PSNR(dB) | Time(sec) |
| --- | --- | --- |
| BM3D[7] | 39.41 | 2.143 |
| WNNM[10] | 38.58 | 331.1462 |
| MLP[11] | 38.26 | 12.786 |
| TNRD[12] | 38.93 | 2.6427 |
| DnCNN[14] | 39.57 | 1.591 |
| Proposed (2-blocks) | **40.26** | **0.7508** |

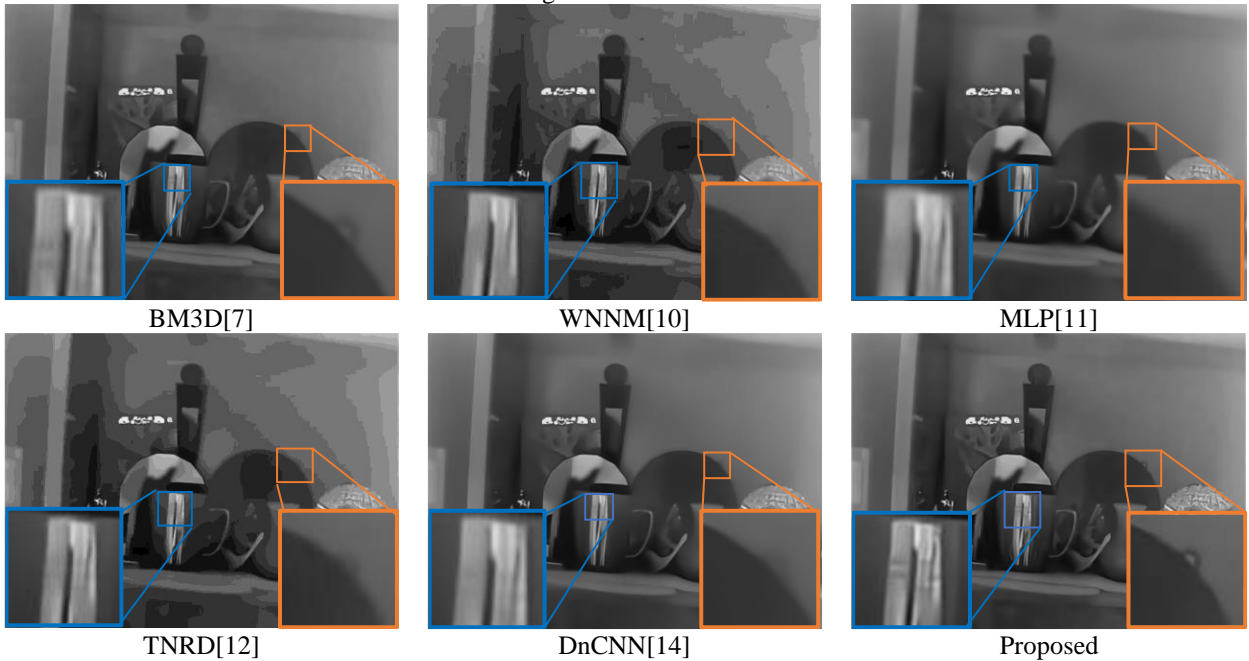

Figure 6: Denoising results with original thermal noise